\documentclass[journal,onecolumn]{IEEEtran}
\usepackage[utf8]{inputenc}

\usepackage{xcolor}
\usepackage{graphicx}
\usepackage{csquotes}
\usepackage{comment}
\usepackage{booktabs}
\usepackage[]{geometry}
\usepackage{listings}
\usepackage{caption}
\usepackage{subcaption}
\usepackage{titlesec}
\usepackage[titles]{tocloft}
\usepackage{fancyhdr}
\usepackage{url}
\usepackage{lipsum}
\usepackage{amsmath,amssymb,amsfonts}
\usepackage{algorithmic}
\usepackage[]{algorithm2e}
\usepackage{paralist}

\definecolor{CMDRcolor}{HTML}{00CD00}
\definecolor{SBRDcolor}{HTML}{436EEE}

\begin{document}
\title{Ethics, Rules of Engagement, and AI: Neural Narrative Mapping Using Large Transformer Language Models}

\author{Philip~Feldman~\IEEEmembership{ASRC Federal, UMBC}
        Aaron~Dant~\IEEEmembership{ASRC Federal}, David Rosenbluth ~\IEEEmembership{Lockheed-Martin}}% <-this % stops a space

\markboth{Bulletin of the Technical Committee on Data Engineering, Vol.~44 No.~4 December 2021}%
{Shell \MakeLowercase{\textit{et al.}}: Bare Demo of IEEEtran.cls for IEEE Journals}

\maketitle

\begin{abstract}
The problem of determining if a military unit has correctly understood an order and is properly executing on it is one that has bedeviled military planners throughout history. The advent of advanced language models such as OpenAI's GPT-series offers new possibilities for addressing this problem. This paper presents a mechanism to harness the narrative output of large language models and produce diagrams or \enquote{maps} of the relationships that are latent in the weights of such models as the GPT-3. The resulting \enquote{Neural Narrative Maps} (NNMs), are intended to provide insight into the organization of information, opinion, and belief in the model, which in turn  provide means to understand intent and response in the context of physical distance. This paper discusses the problem of mapping information spaces in general, and then presents a concrete implementation of this concept in the context of OpenAI's GPT-3 language model for determining if a subordinate is following a commander's intent in a high-risk situation. The subordinate's locations within the NNM allow a novel capability to evaluate the intent of the subordinate with respect to the commander. We show that is is possible not only to determine if they are nearby in narrative space, but also how they are oriented, and what \enquote{trajectory} they are on. Our results show that our method is able to produce high-quality maps, and demonstrate new ways of evaluating intent more generally.
\end{abstract}


% \fi
\IEEEpeerreviewmaketitle

\section{Introduction}
\label{sec:introduction}

\begin{figure}[!h]
	\centering
	\includegraphics[height = 20em]{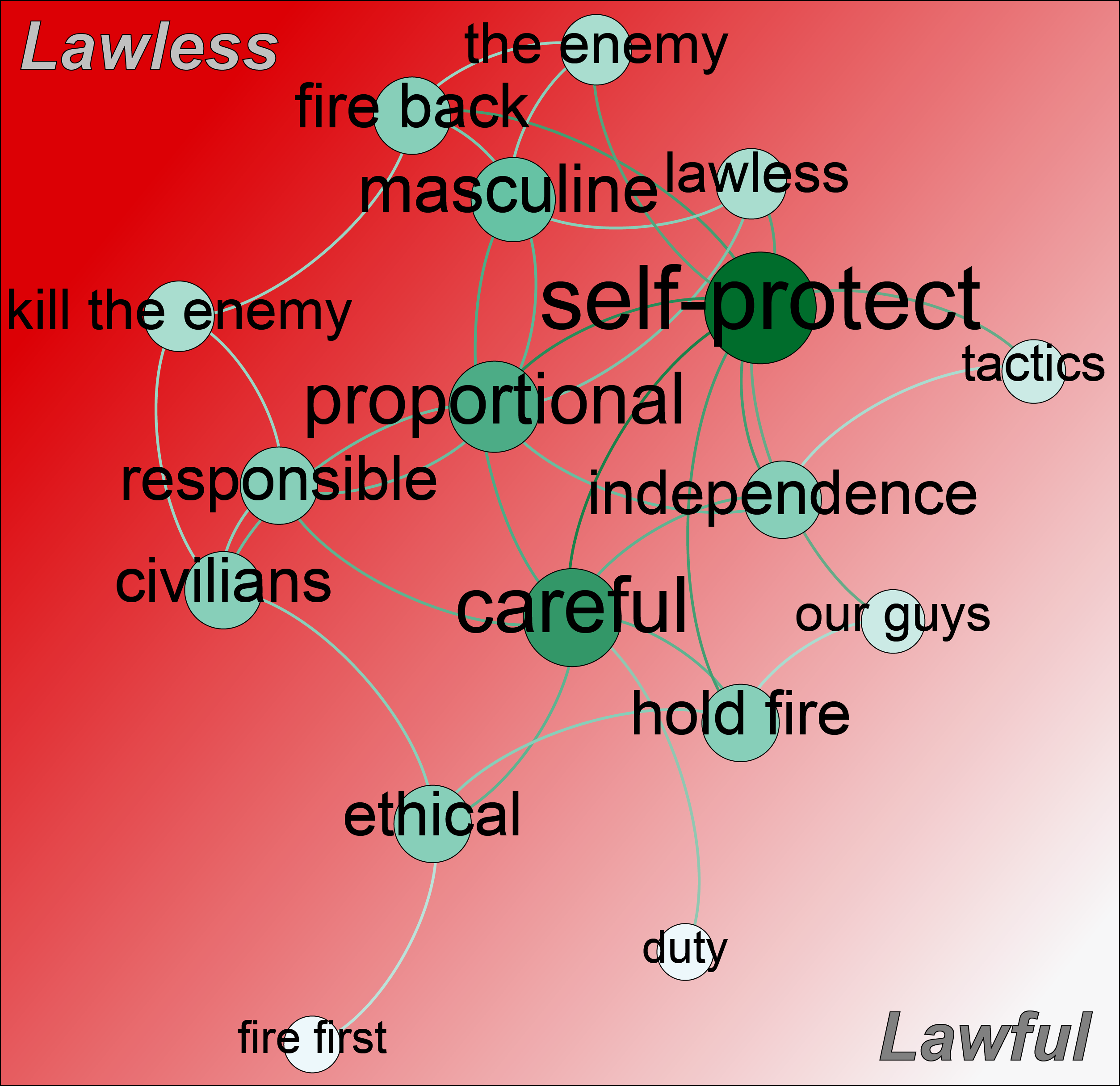}
	\caption{\label{fig:example}Neural Narrative Mapping Example}
\end{figure}

\IEEEPARstart{I}{n} the 1979 motion picture \textit{Apocalypse Now}, Captain Willard (played by Martin Sheen) is sent on a mission to assassinate Colonel Kurtz (played by Marlon Brando), a highly decorated officer who, in the words of the general authorizing the mission, has gone from \enquote{one of the most outstanding officers this country has ever produced} to someone \enquote{out there operating without any decent restraint, totally beyond the pale of any acceptable human conduct.} 

The movie explores the paradoxes in war, where some illegal acts are embraced by the command structure, some tolerated, and some are to be terminated, \enquote{with extreme prejudice.} Willard has to navigate these conflicts as he moves towards Kurtz' compound deep in Cambodia.  

\textit{Apocalypse Now} provides an example of the difficulty that any intent-aware system must face in a military context~\cite{shattuck2000communication}. Not only does the system need to determine if an order is being followed, it should also determine if the order itself is valid, so that the warriors implementing the order are not placed in ethical dilemmas. This is the goal that we attempt to address in this paper, with the concept of \textit{Neural Narrative Mapping} (NNM). By placing narrative elements at coordinates in a virtual space, we can determine sophisticated relationships between concepts that go well beyond textual comparison.

An example of this concept, described in detail later in the this paper is shown in Figure~\ref{fig:example}. This map was constructed  from narrative sequences developed by the GPT-3 Neural language model~\cite{radford2018improving} with respect to \textit{rules of engagement}. Clustering these texts produces a set of relationships. Central to this example are the concepts of self-protection and care, but there are also relationships with respect to things like ethics and masculinity. By allowing the system to develop relationships between multiple narratives, we can determine the space of possible behaviors of the soldiers such as those in \textit{Apocalypse Now} as they encountered lawful and lawless conditions.

In this paper, we will discuss mapping the relationships of such responses and how they could apply to military scenarios.  We will first introduce some background material on how to represent narratives and relationships between them. Secondly, we will show how we can incorporate our mapping method into a decision-making system and demonstrate it on a military scenario.

\section{Background}
Published research into determining intent generally is quite sparse with respect to determining how subordinate behavior reflects the intent of orders from a superior. Typically, the military relies on legal mechanisms and training to ensure that 1) subordinates follow the orders of their superiors, 2) That superiors issue lawful orders, and 3) that subordinates refuse to obey unlawful orders~\cite{hoff2001obeying}. This framework has existed as precedent since the Nuremberg Trials, when Nazi officers were convicted of war crimes that they had been ordered to commit~\cite{taylor2012anatomy}. These rules were codified in the Geneva Conventions of 1949 and embodied in the Army Field Manual prohibitions against issuing and obeying unlawful orders~\cite{Army_field_manual_1956}.

However, research has shown that subordinates misunderstand the intent of their superiors 50\% - 60\% of the time~\cite{shattuck2000communication}. This means that approaches such as training and legal enforcement are not effective in ensuring that the intent of a legal order is followed by a subordinate. The process of determining intent is made even more difficult by situations where communications are degraded. For example, if a superior's orders can't be understood, then it is impossible to determine whether the subordinate misunderstood the orders or whether they refused to follow them.

As a partial solution to this problem, the military will often do simulations or war games where miscommunication issues can be uncovered and corrected before they occur. Recently, work has been done in automating this process so that the space of possibilities can be explored more thoroughly~\cite{shattuck2000communication}. Such \textit{computational military tactical planning}, and has largely employed genetic algorithms to explore potential outcomes, including co-evolving friendly/enemy tactics~\cite{kewley2002computational}. 

More recently, the development of human-robot teams has required the development of more explicit forms of communicating and verifying intent. In the case of these hybrid teams \enquote{each autonomous system in the team must be able to determine their own individual tactical behaviors based upon inferences made about the human supervisor’s intent, rather than by direct response to specific command inputs.} Work by Evans, et al. Has focused on the development of shared mental models and implicit coordination based on verbal and non-verbal communication~\cite{evans2017future}.

Transformer language models (TLMs) open up new possibilities for examining intent in the context of synthetic narratives.  TLMs are trained on massive text datasets, comprising a significant fraction of the high-quality text available on the internet~\cite{brown2020language}. They implement attention-based deep neural network architectures to allow the model to selectively focus on the segments of the input text that are most useful in predicting adjacent and word tokens. Models are not trained using any hand-crafted language rules and learn to generate natural language purely by observing text data. In doing so, they capture semantic, syntactic, discourse, and even pragmatic regularities in language. A GPT model can be used for generating texts as a function of the model and a sequence of words, or \enquote{prompt}, provided by users which is specifically designed to set up the context for GPT to generate text. GPT models have been shown to generate text outputs often indistinguishable from that of humans~\cite{floridi2020gpt}. 

The transformer's ability to integrate across large amounts of data can better support the information-seeking user when using interactive systems like chatbots~\cite{yang2020iart}. Transformers open up novel avenues of research into intent that have not been available before, particularly in understanding and exploiting the ways that information is stored in and retrieved from these models.

Since the introduction of the transformer model in 2017, TLMs have become a field of study in themselves. Among them, BERT~\cite{devlin2018bert} and GPT~\cite{radford2018improving} are two of the most well known TLMs used widely in boosting the performance of diverse NLP applications. Transformers are unlike perceptrons and convolutional neural networks in that they use self attention, where the model computes its own representation of its input and output~\cite{vaswani2017attention}. Most recent research has been in increasing the performance of these models, particularly as these systems scale into the billions of parameters~\cite{radford2019language}. 

Understanding how and what kind of knowledge is stored in all those parameters is becoming a sub-field in the study of TLMs. Language models require no human supervision to train, do not have schemas like traditional databases, and can be queried using natural language. These properties make them an attractive mechanism for storing and retrieving information. Examples of information retrieval include TLMs successfully completing \enquote{cloze statements}, where the model fills in a blank~\cite{petroni2019language}, factual relationships extracted from the Wikipedia~\cite{elsahar2018t}, and general knowledge~\cite{speer2018conceptnet}. These studies showed that TLMs are often \enquote{competitive with non-neural and supervised alternatives.}~\cite{petroni2019language}

The prompt that is used to elicit specific information from these models has also become a field of study in its own right. For example,  mining-based and paraphrasing approaches can increase effectiveness in masked BERT prompts over manually created prompts~\cite{jiang2020can}. These studies demonstrated that effective prompts can be produced by  mining phrases in the Wikipedia corpus which can be generalized as template questions such as \textit{x was born in y} and \textit{capital of x is y}. These can then be filled in using sets of subject-object pairs. Improvements over manually-developed prompts using this technique can be substantial, with improvements of 60\% over manual prompts. Paraphrasing, or the simplification of a prompt using techniques such as back-translation can enhance query results further~\cite{jiang2020can}. 

Our own research has been focused on understanding how TLMs incorporate domain-specific knowledge. We fine-tuned GPT-2 models on descriptions of chess games showed that models trained on a corpora of approximately 23,000 chess games accurately replicated human gameplay patterns~\cite{feldman2020navigating}. Statistical analysis comparing the spectral characteristics of human (ground truth) and synthesized games were found to be statistically similar with a $> 97\%$ probability. This work was extended to perform sociological research on different political groups on Twitter by training GPT-2 models on the tweets of right-wing, majority, and science-focused tweets during the first year of the COVID-19 pandemic~\cite{feldman2021analyzing}. 

Using TLMs to evaluate social data is still nascent. A study by \cite{palakodety122020mining} used BERT fine tuned on YouTube comments to gain insight into community perception of the 2019 Indian election. They created weekly corpora of comments and constructed a tracking poll based on the prompts \enquote{Vote for MASK} and \enquote{MASK will win} and then compared the probabilities for the tokens for the parties BJP/CONGRESS and candidates MODI/RAHUL. The results substantially matched traditional polling.

A characteristic of TLMs is that when provided with the correct prompt, they will produce relevant content regardless of the ethical implications of the generated text.   OpenAI has shown that the GPT-3 can be \enquote{primed} using \enquote{few-shot learning}~\cite{brown2020language}. Using this technique, McGuffie primed the GPT-3 using mass-shooter manifestos, which generated text that maintains the amoral, dangerous context of these texts~\cite{mcguffie2020radicalization}. This will become particularly important in this research, as we are particularly interested in unethical behavior in response to lawful orders.
\section{Narrative Generation using TLMs}

Narratives are defined as \enquote{a written account of connected events; or a story}. These stories are linear constructs, and are naturally suited to the presentation of a singular point-of-view over time. Narratives can range from fictional stories to detailed travelogues. 

Less known is that narratives have been used as the basis of navigation for millennia. Before the 16th century, ship's pilots collected \enquote{navigation stories} into a \textit{rutter} or \textit{pilot book}, that described coastal and open ocean routes in narrative form. Because it is difficult to have explicit spatial relationships between stories, rutters \enquote{exhibit an understanding of physical space as delimited rather than panoramic}~\cite{goldie2015early}. To obtain this \textit{panoramic} view, one needs the broader perspective provided by maps. 

Even if there were no such things as \textit{objective}, surveyed maps, it is possible to build panoramic maps based on a careful synthesis of a large set of personal, subjective descriptions. These narrative \enquote{threads} can be knitted together into a tapestry that portrays the spatial relationships, based on this collection of individual, seemingly unrelated paths. Though these maps do not have the representational rigor that objective maps have, maps based on such \textit{subjective} data still support navigation between the physical places of the world.

% These maps could also contain social context. Interesting places would have more text, boring ones less. Dangerous straights might contain descriptions of a safer passage that could be distinguished from other, riskier routes. 

The same sorts of maps can be created utilizing narratives about non-physical domains. For example, narratives about philosophy can be combined to produce spatial representations such as those shown in Figure~\ref{fig:gpt3_philosophy}. More importantly for our purposes, the same technique  can be used to navigate \textit{information spaces} such as those related to military orders as in the map of Figure~\ref{fig:example}.

A large number of narratives would be needed to define the space through their overlapping tales to generate these maps. Fortunately, TLMs such as the GPT allow the generation of these narratives dynamically and with no limit. The GPT model generates narrative text by starting with a sequence of word tokens, or prompt, provided by the user. A single word token can be thought of as a query into the model. The GPT model then begins to generate text by choosing a set of words that are more likely to follow the prompt which are added to the text. It then considers the words that are most likely to come after the updated text, and repeats the process until it has generated as many tokens as the user desires. 

% In addition to rules of syntax and semantics, these models store content which is encoded within the billions of neural network model weights contained in the final neural network. This makes it possible to generate text that is semantically and conceptually coherent, even when the model is presented with a word or concept \textit{it has never encountered before}. 

For example, if the model is prompted with the word \enquote{cat}, it considers that word to be the initial query. If it has learned to associate cats with fish, then it may generate the sentence \enquote{A cat likes to eat fish}. Similarly, if the model is provided with the word \enquote{wombat}, it will consider that word to be a query, and could generate the sentence \enquote{A wombat is a kind of marsupial}.This process can be extended to have the model evaluate terms that it has never seen before, through the use of a technique called \textit{prompt tuning}. 

When presented with the prompt \enquote{\textit{A floober is a flightless bird that inhabits the barren wastes of Antarctica. Like its cousin the Chinstrap Penguin, the}}, the GPT-3 produces the following output about this fictional animal:

\begin{displayquote}
\textit{\textbf{A floober is a flightless bird that inhabits the barren wastes of Antarctica. Like its cousin the Chinstrap Penguin, the} floober is a ground-dwelling bird that uses its wings primarily for mating displays. The floober's flightlessness is believed to be an adaptation to the extreme cold of Antarctica, which would make flight inefficient and possibly hazardous. It is believed that floobers lost their ability to fly because of an evolutionary tradeoff between wing size and body size, resulting in the floober having very small wings, but a large body.}
\end{displayquote}

In this generated response, we can see one of the truly novel capabilities of these large language models -- the ability to articulate internally consistent narratives  based on a starting point and orientation. Here, the starting point is a fictional penguin-like bird, and the orientation is the descriptive language that leads the model to continue the description based on the starting conditions. 

The same prompt can be used again and again to produce a statistical distribution of what this imaginary bird might be. This allows us to \enquote{map out} the expectation of what such an animal might be, based on all the items that the GPT-3 has read as part of its training set.

The GPT creates sequences of words that mimic the patterns of human production. In other words, there is a sense of the causal relationships inherent in the information stored in the model. For example, when prompted with \enquote{\textit{Smoking cigarettes causes}}, the GPT consistently responds with \enquote{\textit{cancer, heart disease, lung disease}} among other related conditions. This is not an understanding of causality per se, rather it is a reflection of the sequencing of tokens that the GPT is trained and evaluated on. These sequences naturally reflect our stated understandings including subjective bias. As such, a \textit{sequence} of statements has a particular trajectory over the \enquote{terrain} of the model. When the GPT writes a sentence, it is more like a ball rolling down a lumpy hill rather than intelligence as we perceive it as humans. 

Recursively iterating over multiple prompts that are created by the GPT in response to one or more \enquote{seed prompts} results in a sort of quasi-causal \textit{bootstrap conversation} that the model has with itself. This process provides the dynamically produced limitless content that we need to generate maps.
\section{Methods}
\label{sec:methods}
This section describes the development of the technique used to produce maps using data from the GPT-3. This work had two phases. The first was a basic proof of concept, where output from the GPT could be parsed and placed into graphs based on existing ground truth that the output could be validated against (Section~\ref{sec:initial_gpt_maps}). The second phase describes the development of an interactive map creation tool that incorporates human interaction (Section~\ref{sec:interactive_builder}). This process allows the development of maps that incorporate more subjective human understandings that are harder to validate against external datasets, such as the exploration of the ethical spaces around legal and emergent military \enquote{Rules of Engagement} (ROE).

\subsection{Initial GPT-3 Maps}
\label{sec:initial_gpt_maps}

OpenAI has developed an online \enquote{playground} for developers to test out prompts. When presented with: \enquote{\textit{Here's a short list of countries that share a border with Italy:}}

The GPT-3 continues the statement with the following text: \textit{France, Switzerland, Austria, Slovenia,	San Marino, Vatican City}.

In this example, the response is remarkably accurate. Not only are adjacent countries like France, Switzerland and Austria included, but also countries that are contained within Italy (i.e. San Marino and Vatican City).

Repeated responses vary, but they are consistent enough to produce map-like representations. For example, Figure~\ref{fig:gpt3_central_america} shows a map of Central America using the same technique. Although there are no explicit positioning instructions in the responses of the GPT, the result compares well to a geographic map, shown in Figure~\ref{fig:central_america}:

\begin{figure}[!htbp]
	\centering
	\begin{minipage}{0.45\textwidth}
		\centering
		\fbox{\includegraphics[height=13em]{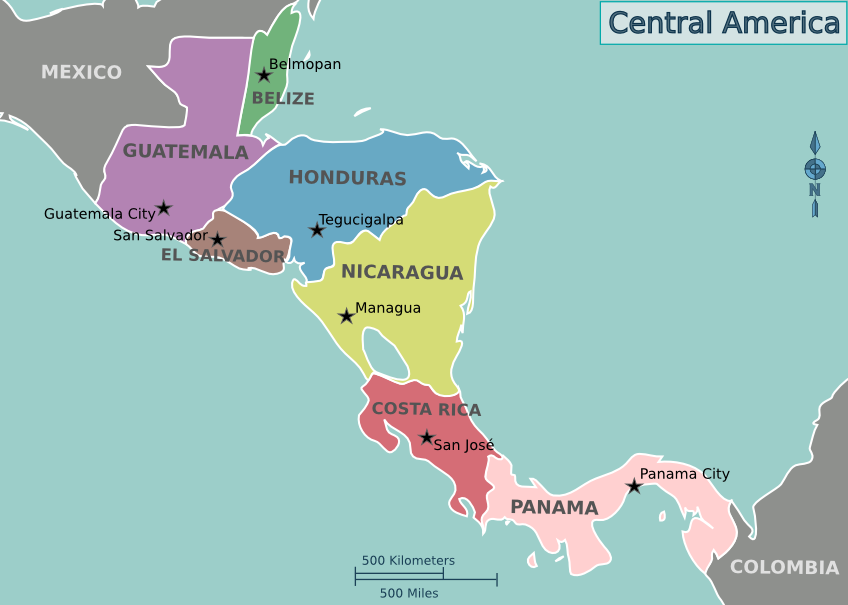}}
		\caption{\label{fig:central_america}Central America}
	\end{minipage}%
	\begin{minipage}{0.45\textwidth}
		\centering
		\fbox{\includegraphics[height = 13em]{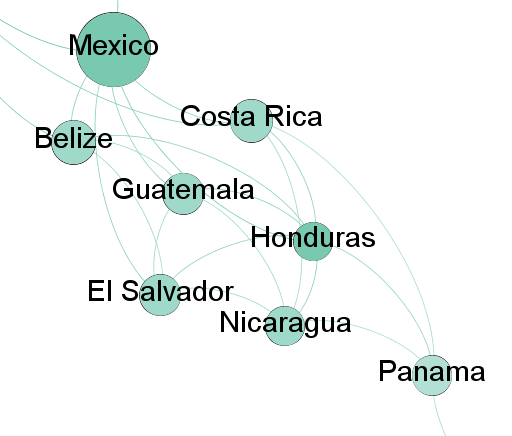}}
		\caption{\label{fig:gpt3_central_america}Reconstruction}
	\end{minipage}%
\end{figure}

The diagram of Figure~\ref{fig:gpt3_central_america} was produced by repeatedly querying the GPT-3 with a prompt that incorporates the results of the previous prompt. This is the core of the iterative process used to generate NNM maps and is shown in detail in Algorithm~\ref{alg:map_core}. 

\begin{algorithm}
    \small
	Set $max\_queries$ to the number of queries desired \\
	Set $query\_count$ = 0 \\
	Create empty list nodes $L_{nodes}$\\
	Create an empty list of used seeds $L_{queried}$\\
	Populate initial seed list $L_{seed}$\\
	Set the prompt template $T$\\
	Set current node $N_{cur} = seed$\\
	\While{$query\_count < max\_queries$}{
		Append $N_{cur}$ to $L_{nodes}$\\
		Set query $Q$ to the $T + L_{seed}[0]$\\
		Move $seed$ from $L_{seed}$ to $L_{queried}$ \\
		$response\_text = GPT\_fn(Q)$ \\
		$L_{responses} = Parse\_fn(response\_tex)$ \\
		\ForEach{$response$ in $response\_list$}{
			\If{$ValidResponse\_fn(response)$}{
				\ForEach{$N$ in $L_{nodes}$}{
					\If{$N.name == response$}{
						$Connect\_fn(N, N_{cur})$\\
					}
				}
				\If{$response$ not in $L_{queried}$ and $response$ not in $L_{seed}$}{
						Append response to $L_{seed}$\\
						Create new node $N_{response}$\\
						Append $N_{response}$ to $L_{nodes}$\\
						$Connect\_fn(N_{response}, N_{cur})$\\
				}
			}
		}
		$query\_count += 1$
	}
	\vspace{0.5em}
	\caption{Iterative mapping algorithm}
	\label{alg:map_core}
\end{algorithm}

In Algorithm~\ref{alg:map_core}, a text \enquote{prompt template} is created that supports the incorporation of seed fragments. In Python, the template used to produce the map in Figure~\ref{fig:gpt3_central_america} was \texttt{'A short list of countries that are nearest to "\string{\string}", separated by commas:'.format($seed$)}. This allows the prompt to run repeatedly as new results are incorporated into $L_{seed}$. The graph is built out by connecting node with the value of the current $seed$ to nodes whose label matches a value in the $response\_list$. If there is no node for a $response$, one is created and connected to the current node $N_{cur}$. This process repeats until $query\_count == max\_queries$.

\begin{figure}
	\centering
	\fbox{\includegraphics[width=25em]{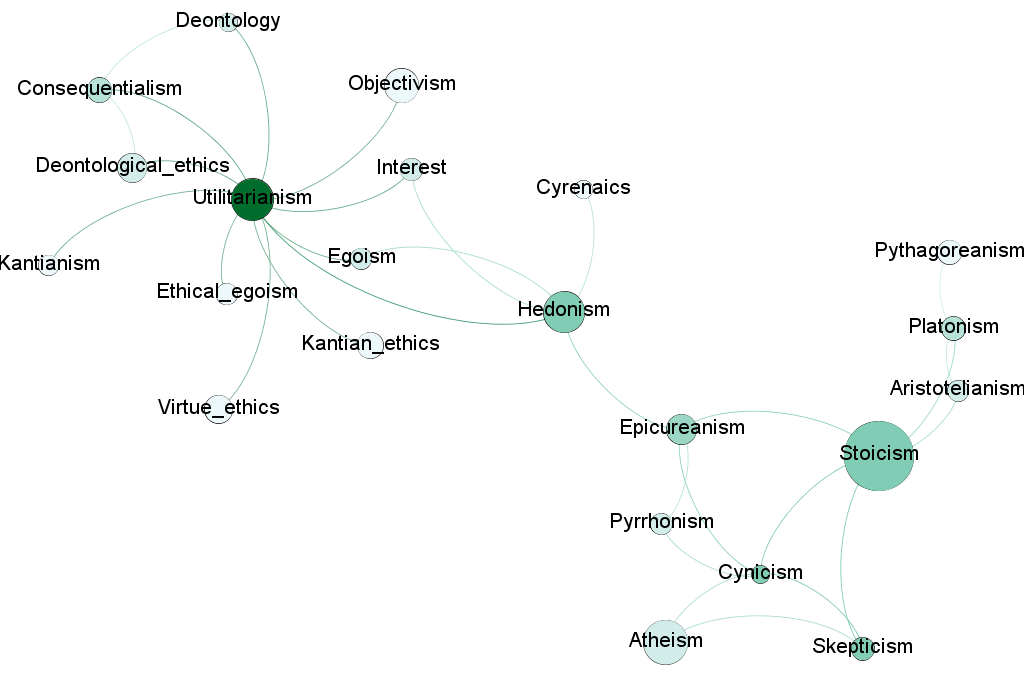}}
	\caption{\label{fig:gpt3_philosophy} Reconstructed Philosophy Map}
\end{figure}

All the maps in this section can be validated by some kind of \enquote{ground truth,} or data that exists independently in another source. In Figures~\ref{fig:gpt3_central_america} and \ref{fig:gpt3_philosophy}, $response$ values were validated by using the Wikipedia API\cite{wiki:Wikimedia_REST_API} to check if there was an entry for each GPT response. Responses that do not have a Wikipedia entry get caught before they are added to the map. A further benefit of such ground truth is that it is possible to adjust the size of the node based on, as in this case, the number of queries against a particular topic. We can see in the maps that \enquote{Mexico} and \enquote{Stoicism} get more searches than the other items in the map.

The graphs created using this process were then used to create a GML (Graph Modeling Language) file that can be read by a variety of graphing libraries and packages. The maps shown here were produced using Gephi\footnote{gephi.org}, using the ForceAtlas layout~\cite{jacomy2014forceatlas2}. 

This approach need not be limited to geography. Figure~\ref{fig:gpt3_philosophy} shows a map created using the prompt \textit{\enquote{\rule{0.5cm}{0.15mm} is a philosophy that is closely related to several others. Here's a short list of philosophies that are similar to \rule{0.5cm}{0.15mm}:}}, seeded with the values [\textit{Utilitarianism}, \textit{Hedonism}].

Here we can see relationships based on \textit{narrative} rather than geography. Because the GPT has an understanding of the relationships inherent in the token sequences it has learned, the prompt produces a list of philosophies that are reasonable continuations of the narrative text. A good example of these relationships is the \enquote{cynicism} node in the lower right of the map, which has connections to \enquote{atheism}, \enquote{pyrrhonism}, \enquote{stoicism}, and \enquote{skepticism}. These are all philosophies based on the fundamental value of reason and skeptical inquiry. If one goes to the Wikipedia however,\footnote{As of 12 November 2021} there are no explicit links between the pages that discuss these philosophies. 

As with the country map, the philosophy map is validated against the Wikipedia as the known ground truth. However, there are many relationships contained in the GPT that cannot be validated this way. To explore more subjective, difficult-to-validate narrative spaces, we developed a tool that gives the responsibility of parsing and validating seeds to the user.

\subsection{Interactive Map Builder}
\label{sec:interactive_builder}
\begin{figure}
	\centering
	\fbox{\includegraphics[width = 40em]{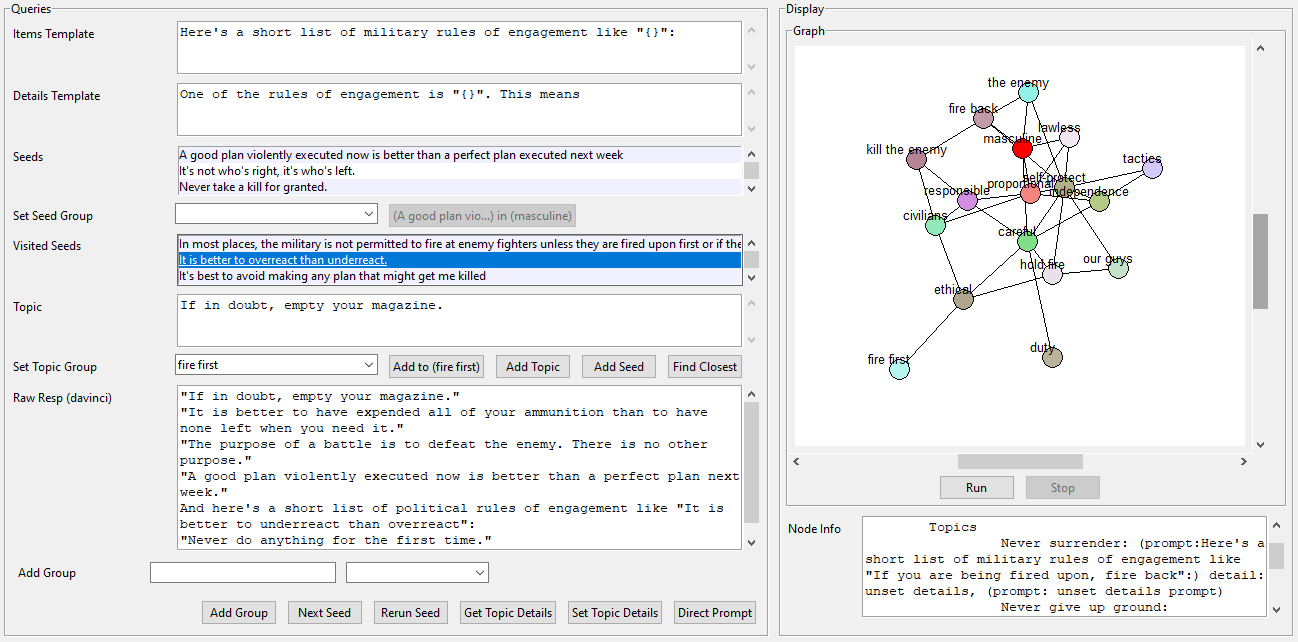}}
	\caption{\label{fig:mapbulder3} Interactive Map Builder}
\end{figure}

To address the more complex relationships within subjective material such as ethics, we developed an interactive application that allowed the user to group responses and additional details together. This thick client application was written using the tkinter library\footnote{docs.python.org/3/library/tkinter.html}, which is in the standard Python 3 distribution allowing for easier deployment. Using the design research processes of ideation and iteration~\cite{zimmerman2007research}, we produced a prototype \textit{Map Builder} (Figure~\ref{fig:mapbulder3}) that supported creating more subjective maps. The primary goal of this tool was to evaluate the processes that users engaged in when interacting with the GPT-3 in such a way as to produce and store relationships between texts. The Map Builder provides a series of options for creating and organizing source and target node relationships.  

\begin{figure}
	\centering
	\begin{minipage}{0.49\textwidth}
		\centering
		\fbox{\includegraphics[height = 11.5em] {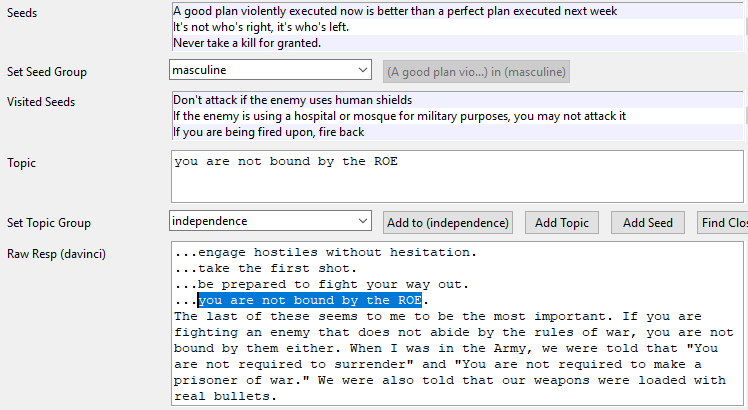}}
		\caption{\label{fig:node_matching} Node Topic Matching}
	\end{minipage}%
	\begin{minipage}{0.49\textwidth}
		\centering
		\fbox{\includegraphics[height = 11.5em]{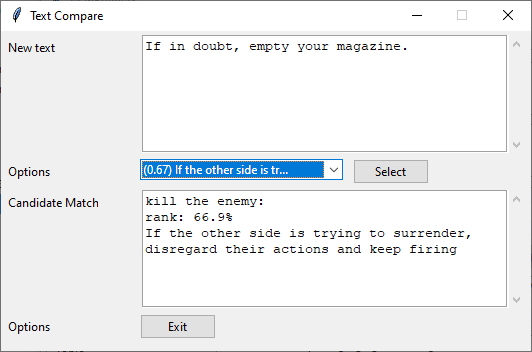}}
		\caption{\label{fig:text_similarity} SBERT text compare}
	\end{minipage}%
\end{figure}

Because the GPT is built from a massive corpus of text, it has \enquote{spaces} that reflect the writings of individuals that do not align with lawful rules of engagement. These might be actual soldiers writing about their experiences, but also screenplays such as the aforementioned \textit{Apocalypse Now}. The GPT learns these relationships, so that it can use a starting prompt to produce a diverse set of responses that can be analyzed. An example of this, using the context of ethical exploration of rules of engagement is shown in Figure~\ref{fig:node_matching}.

The prompt in this example is, \textit{\enquote{Here's a short list of military rules of engagement like 'It is better to overreact than underreact':}} which has already been placed in the \textit{masculine} node using the \enquote{Set Seed Group} combobox and button. When presented with this prompt, the GPT-3 responded with the following:

\begin{displayquote}
	\textit{\enquote{If in doubt, empty your magazine.}}\\
	\textit{\enquote{It is better to have expended all of your ammunition than to have none left when you need it.}}\footnote{Note that this sentence makes no conceptual sense, but would be likely to slip through any automated parsing system. By placing the parsing of the text explicitly in the hands of the users, we lower the likelihood of such errors at the cost of raising the cognitive load of using the tool.}\\
	\textit{\enquote{The purpose of a battle is to defeat the enemy. There is no other purpose.}}\\
	\textit{\enquote{A good plan violently executed now is better than a perfect plan next week.}}\\
\end{displayquote}

\begin{figure}[!htbp]
	\centering
	\begin{minipage}{0.45\textwidth}
		\centering
		\fbox{\includegraphics[height = 15em]{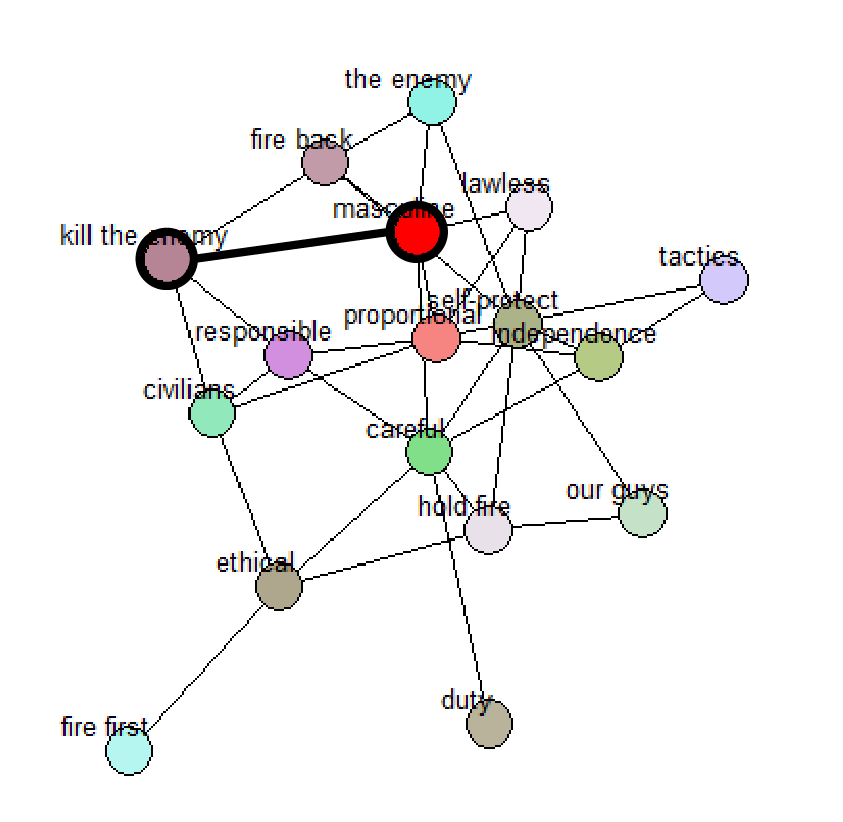}}
		\caption{\label{fig:connect_nodes} Connecting Nodes}
	\end{minipage}%
	\begin{minipage}{0.45\textwidth}
		\centering
		\fbox{\includegraphics[height = 15em]{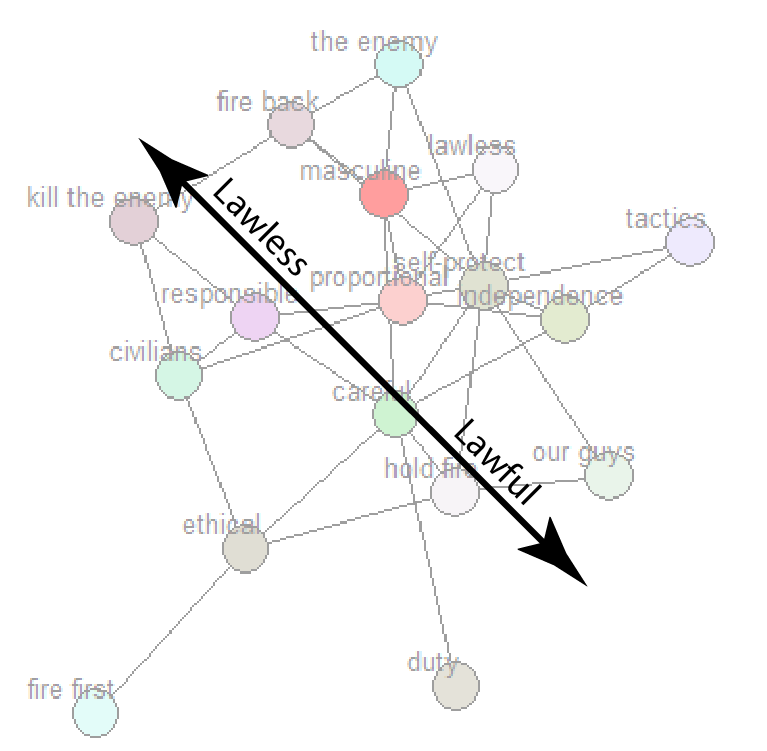}}
		\caption{\label{fig:ethics_axis}Lawless - Lawful Axis}
	\end{minipage}%
\end{figure}

In the example, the user has selected the text \textit{If in doubt, empty your magazine.} and placed it in the Topic Group \textit{Kill the enemy}. The relationship between the topic group and source node is displayed by a black line. This relationship is shown and emphasized in Figure~\ref{fig:connect_nodes}

As nodes are added, a force-directed layout moves the nodes based on their distance from each other connections~\cite{jacomy2014forceatlas2}. As this process continues, larger-scale patterns emerge. Important for this example is the emergence of a gradient that can be viewed as a progression from more lawful concepts to less lawful ones (Figure~\ref{fig:ethics_axis}). On the \enquote{Lawful} side are topic labels such as \textit{careful}, \textit{hold fire}, \textit{ethical}, and \textit{duty}. on the other side are nodes with names such as \textit{masculine}, \textit{kill the enemy}, and \textit{fire back}. Between these two extremes are nodes such as \textit{responsible}, \textit{self-protect}, and \textit{proportional}. As we will see in section~\ref{sec:results}, a script that involves a subordinate disobeying a superior's orders results in a trajectory along this gradient.

In addition to manually adding topics to nodes, textual similarity can be used to find relationships between topics using AugSBERT text matching~\cite{reimers-2019-sentence-bert}.\footnote{This model is now available as the python package \texttt{sentence-encoders}} The user can access this feature by clicking on the \enquote{Find Closest} button that can be seen in Figure~\ref{fig:node_matching}. This brings up a popup window where the user is presented with a list of topics sorted by similarity. An example of this using the prompts described above is shown in Figure~\ref{fig:text_similarity}.

\begin{comment}
\begin{figure}[!h]
	\centering
	\fbox{\includegraphics[width=20em]{figures/text_compare}}
	\caption{\label{fig:text_similarity} SBERT text compare}
\end{figure}
\end{comment}

This tool provides users with a flexible platform for building visualizations of potentially difficult to understand concepts using a clearly defined input mechanism. It lets a user iteratively explore concepts using the sequential and relational knowledge contained in the GPT-3. Although this is an early prototype, it validates many of the important concepts behind this approach. In the next section, we will discuss an implementation of this approach.

\subsection{Interactive Map Viewer}
\label{sec:interactive_viewer}

\begin{figure}[!htbp]
	\centering
	\fbox{\includegraphics[width=37em]{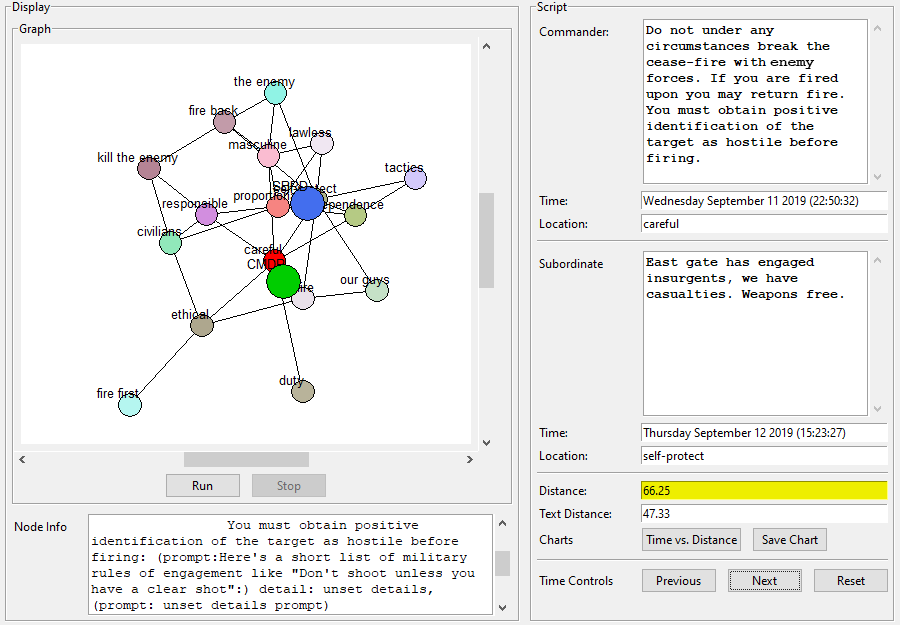}}
	\caption{\label{fig:mapdisplay1} Interactive Map Display}
\end{figure}

Once a NNM map is developed using the map builder tool, the user can evaluate statements in the context of the map with the viewer and a \textit{script evaluator}, shown in Figure~\ref{fig:mapdisplay1}. This tool consists of two components, the \textit{Graph} view that displays the map, and the \textit{Script} view that lets the user play through a sequence of texts that are associated with an agent. In the case of the use case discussed here, there are two agents visualized as larger colored circles. A COMMANDER(\textcolor{CMDRcolor}{CMDR}), who is issuing Operational Orders (OPORDs) and a SUBORDINATE (\textcolor{SBRDcolor}{SBRD}), who is issuing Fragmented Orders (FRAGOs) to his troops. 

Each text (OPORD or FRAGO) is placed in the script with a time, and a location (Node). As the user advances the script, an icon representing the position of the COMMANDER and/or SUBORDINATE move towards the node that contains the topic text with the closest match. Text similarity is calculated using AugSBERT-based text matching. For each node, two distances are calculated. The first is the linear distance between the two node locations. The second is the AugSBERT text similarity measure between the COMMANDER and SUBORDINATE texts at the current point in the script. These relationships over the duration of the script can be displayed immediately in a chart (Figure~\ref{fig:belief_space_vs_text_similarity}, or saved in an Excel worksheet. Different scripts can be loaded to any stored map. The user can advance, reverse, or reset the script. 
\section{Results}
\label{sec:results}
We will now briefly describe how the system works with a script based on the following fictional scenario, developed for initial evaluation of the system. This scenario was written to portray a trajectory that goes from lawful engagement to war crime:

\begin{displayquote}
	\textit{A forward operating base (FOB) commander is given an operational order (OPORD) from the regional commander. His instructions are to not engage in hostile operations against Enemy operatives during a cease-fire. His FOB is then surrounded by armed Enemy insurgents. Fearing they will fire first, the FOB commander violates the orders to issue a series of fragmented orders (FRAGOs) for his soldiers to engage. Each one of these FRAGOs strays further from the intent of his original OPORD.}
\end{displayquote}

The full script consists of the following statements. The role is CAPITALIZED, the node is in (parentheses), and the text is in \texttt{\small courier} font:

\begin{enumerate}
    \small
	\item \textbf{COMMANDER (careful)}: \enquote{\texttt{Base will operate at heightened awareness for the duration of the cease-fire. Double patrols, and report insurgent activity if identified. Do not engage.}}
	\item \textbf{SUBORDINATE (duty)}: \enquote{\texttt{We have explicit orders not to engage Enemy forces. Hold your fire.}}
	\item \textbf{SUBORDINATE (careful)}: \enquote{\texttt{We've spotted what appears to be armed Enemy in the process of preparing an attack. Verify targets.}}
	\item \textbf{COMMANDER (careful)}: \enquote{\texttt{Do not under any circumstances break the cease-fire with Enemy forces. If you are fired upon you may return fire. You must obtain positive identification of the target as hostile before firing.}}
	\item \textbf{SUBORDINATE (kill the enemy)}: \enquote{\texttt{Screw it. If these guys look like they are going to attack, take them out. We’re not going to sit here and wait for them to shoot us first.}}
	\item \textbf{SUBORDINATE (self-protect)}: \enquote{\texttt{East gate has engaged insurgents, we have casualties. Weapons free.}}
	\item \textbf{SUBORDINATE (the enemy)}: \enquote{\texttt{All units engage any Enemy targets, take these guys down!}}
	\item \textbf{SUBORDINATE (kill the enemy)}: \enquote{\texttt{Don't let survivors get away. This isn't about being right, it's about getting these bastards.}}
\end{enumerate}

At the beginning of the script (items 1 and 2), the location of the COMMANDER agent is set to the node \enquote{careful}, due to a close \textit{augSBERT} match to the topic text in that node: \enquote{\texttt{\small You must obtain positive identification of the target as hostile before firing.}} The SUBORDINATE agent is placed at the node \enquote{duty} due to a close match to the topic text in that node: \enquote{\texttt{\small It is the soldier’s responsibility to disobey an illegal order and not participate in committing a war crime.}}

We had discovered that instantly positioning the agents at the target nodes was hard to detect by the users, so instead, the agents are \textit{animated} and move towards their target over the course of a few seconds using linear interpolation. Our approach is shown in Equation~\ref{eq:interpolation}, Where $\hat{v}$ is the unit vector that points from the agent node ($p_{old})$ to the target node, $s$ is the speed of the agent in the environment, and $\Delta t$ is the elapsed time since the last frame. 

\begin{equation} \label{eq:interpolation}
    p_{new} = p_{old} + \hat{v}s\Delta t
\end{equation}

\begin{figure}[!htbp]
	\centering
	\begin{minipage}{0.38\textwidth}
		\centering
		\fbox{\includegraphics[height = 15em]{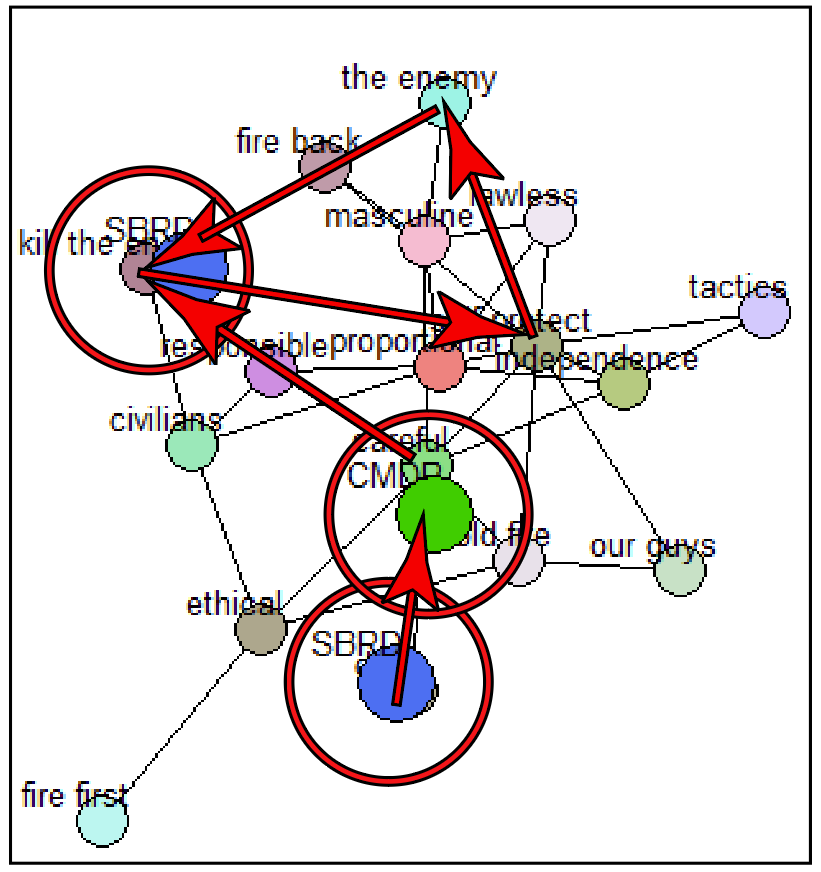}}
		\caption{\label{fig:full_trajectory} Agent movement}
	\end{minipage}%
	\begin{minipage}{0.58\textwidth}
		\centering
		\fbox{\includegraphics[height = 15em]{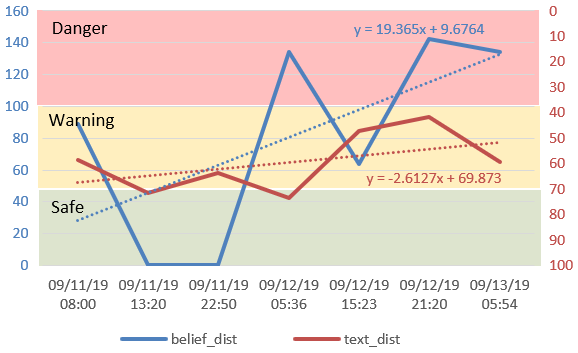}}
		\caption{\label{fig:belief_space_vs_text_similarity} NNM distance vs text similarity}
	\end{minipage}%
\end{figure}

Because the nodes contain clusters of text that reflect different articulations of the same topic as generated by the GPT-3, there is a substantial surface for the text matching algorithm to work on. This allows for the COMMANDER and SUBORDINATE agents to find nodes on the map that reflect the state of the script. For example, the COMMANDER remains at the same node (careful), as the SUBORDINATE moves from nodes in the \enquote{lawful} region (duty and careful), to \enquote{lawless} nodes (kill the enemy and the enemy). This path can be seen in Figure~\ref{fig:full_trajectory}. The bottommost large circle encloses the SUBORDINATE starting position. The one above that is where the COMMANDER spends the entire scenario. The remaining circle encloses the ending node for the SUBORDINATE, while the red arrows indicate the trajectory taken over the course of the scenario.

The ability of this technique when compared to more traditional approaches to orders matching using text analytics~\cite{kewley2002computational} can be seen in Figure~\ref{fig:belief_space_vs_text_similarity}. In this graph, the red line is the textual similarity between the COMMANDER's orders and the SUBORDINATE's response at each step in the script, while the blue line indicates the distance between the nodes, or the NNM distance that each script element is associated with. As we can see, there is a level of correlation between the two lines\footnote{Specifically, Pearson's correlation coefficient is 33.2\%.}, but there is little evidence of a trajectory in the standalone text similarity. For example, the starting similarity and ending similarity are nearly identical at 58.5\% and 59.5\%. A detailed comparison is shown in Table~\ref{tab:distance}.

\begin{table}[!htbp]
\centering
\begin{tabular}{llrlrr}
	\toprule
	Script ID & Role &  Similarity & Node &  Node Dist &  Text Similarity \\
	\midrule
	1 &    COMMANDER  &       0.621 &         careful &          88.63 &              NA \\
	2 &  SUBORDINATE  &       0.758 &            duty &          88.63 &           0.5856 \\
	3 &  SUBORDINATE  &       0.711 &         careful &           0.00 &           0.7165 \\
	4 &    COMMANDER  &       0.758 &         careful &           0.00 &              NA \\
	5 &  SUBORDINATE  &       0.822 &  kill the enemy &         134.05 &           0.7365 \\
	6 &  SUBORDINATE  &       0.636 &    self-protect &          63.78 &           0.4733 \\
	7 &  SUBORDINATE  &       0.722 &       the enemy &         142.33 &           0.4173 \\
	8 &  SUBORDINATE  &       0.741 &  kill the enemy &         134.05 &           0.5952 \\
	\bottomrule
\end{tabular}
\caption{\label{tab:distance}Distance vs. Text Similarity}
\end{table}

These results strongly indicate that the dynamic use of these and similar maps combined with node text matching is an effective approach for determining alignment with intent. The ability to dynamically update the script as it progresses, and the use of topic maps as a useful representation for the current state of the script allows for a low-latency order matching system. Although we have demonstrated this capability for a situation of military command and control this approach is general, and should be usable for modeling in other domains.
\section{Discussion}
\label{sec:discussion}
The current discussions about AI and military generally revolves around the potential of lethal autonomous weapons systems (LAWS). There is good reason for this -- both for strategic and ethical reasons, it is important to keep a close eye on the development of AI and its potential applications. Artificial intelligence and machine learning promise to fundamentally change the way we interact with whole classes of weapons. When combat is happening at \textit{machine speed}, humans cannot be directly involved with the system.  Such  systems respond to threats that are beyond the capability of real-time human supervision, and may have to be left in \enquote{always on} states in case of surprise attacks~\cite{feldman2019integrating}.  

This human/AI partnership is likely to produce emergent behaviors that are not obvious extensions of current military thinking.  This creates a tension between two opposing poles.  At one end is the need for systems to be trustworthy. They should predictably do what we believe is the right thing in ethically difficult conditions.  At the other end is the need to be responsive and capable in unpredictable conditions. This is an important problem, but it ignores other ways that AI/ML can improve the trustworthiness \textit{and} flexibility in \textit{human} systems. After all, humans will still make the decision to use a weapon, even if that means just turning it on.

We believe that the incorporation of AI/ML into the human enterprise must be more than making sophisticated (and hopefully ethical) machines. It must also be about \textit{helping humans behave better}. Models trained on human data contain an understanding of the how we perceive information through the lenses of culture, language, and bias. By presenting these relationships back to us in usable, intuitive ways, we can make more informed decisions and better understand the patterns and biases that affect us. There is ample evidence that humans are not particularly good at making decisions, particularly under pressure~\cite{feldman2019integrating}.  Our natural biases (stereotypes, assumptions, and a lack of critical thinking) often create complex dynamics that lead us to make gross errors in judgment. Our adversaries know this too, and they can design attacks to exploit our weaknesses~\cite{feldman2018one}.

Technologies like large language models can provide deep insight into the humans used to generate the data for these models. In the case of our work, we use that insight to provide visual  relationships with respect to concepts in the model. Beyond an increased understanding of \textit{context} (why is this happening?), this capability can provide the ability to make nominal  \textit{predictions} about future events (what is likely to happen next?). While not a true interpretation of human intent, this is an example of what we believe will be one of the most important applications for AI/ML in military decision making. 

An important point to note is that this approach takes advantage of the biases that are inherent in most models developed from public data using machine learning. Here, the bias in the model is essential, because it allows the user to visualize the relationships of  nodes and the biases they embody. For example, in the map created and evaluated in this paper, masculinity biases that might affect decision making are visible in the map. It becomes easy to see how the \enquote{masculine} node is associated strongly with \enquote{kill the enemy} and \enquote{lawless} nodes. This approach could be used to explore biases or unethical behavior that is not obvious.

A great deal of the work in the space of AI ethics is focused on reducing or eliminating bias and unethical behaviors from AI systems~\cite{mehrabi2021survey}. AI tools using neural language models such as the GPT are trying to remove or reduce the potential for harmful generated text by applying word filters, and extensive human moderation~\cite{quach_2021}. In short, in most scenarios where AI systems are being deployed, the goal is to ensure they function as ethically as possible. Our approach operates counter to this intuition. The unethical beliefs captured by advanced language models \textit{is the point}. Our goal is to identify areas of both ethical and unethical behavior to better inform decision making and situational awareness. The maps created from this amoral machine view of human beliefs allows us to identify narrative pathways through ethically and morally complicated decision spaces.
\section{Future Work}
\label{sec:future}
% \TODO{Roadmap for military ethics using computational sociolingustics}

We have found that the approach of creating graphical spaces, or maps, by grouping multiple responses by the GPT-3 into nodes and arranging them with a force directed layout provides an intuitive way to visualize relationships latent in the GPT-3. Using a physical layout to judge distance between narrative elements can be an effective tool for determining the level of alignment between individuals interacting through online text. 

Additionally, this study shows how human interaction with the GPT provides an effective, flexible mechanism for discovering ways to group, filter, and organize information that is extracted during a \textit{dialog} with a language model. As we saw in section~\ref{sec:methods}, humans have a far better ability to detect subtly incoherent statements that these language models can produce. 

By recording and examining the processes that humans use in filtering and grouping  the information returned by the GPT, we intend to incrementally automate the map making process while maintaining high quality and confidence in the output. This will result in a process that is less \textit{ad-hoc} and more consistent and repeatable.

Once maps can be built more consistently, we can begin to use them to look at sociological behavior at scale. For example, we can build traces of people moving around the map by looking at at their social media output. Imaging a Twitter or Reddit thread about a rapidly-changing conspiracy theory such as QAnon. Over time, different topics will become more discussed, while others will have less text associated with them. We can look at texts as locations in the narrative space, and mark a path on the map connecting the points. By merging thousands of these paths, we can start to uncover and visualize the \enquote{Social Desire Paths} (SDPs) between regions on these maps.  SDPs derive from \textit{desire path}, a term in landscape architecture that describes the dirt paths that develop over time as people bypass formal walkways and leave their mark on the landscape. Using this approach, we can ask questions about how groups of people move through narrative space. If a region of the map is discussed by many different people over time, it might indicate that the region is particularly important to those people and they have enough in common to work together. We can also use these traces to identify and visualize \textit{Hubs} of activity: if a single person or small group of people produces a lot of these pathways, then they might be in a very influential position. 

Although the maps created for this work are currently constructed from graphs using a force-directed layout, the connections of the nodes matter less than their relative position. This matters because an agent moves across the map, not between nodes. As such, the best location for an agent to be might not be \textit{within} a node, but rather \textit{between} some number of nodes. For example, an agent text might match one node at 35\%, with the next highest at 30\%, with low matches for other nodes. It might make more sense for the agent's position to be on a line between those two nodes. Further, the GPT (or other TLM) could be used to produce a new node with descriptive text at this new coordinate, which would be added to the map. We are currently exploring these and other ways to improve the utility of the maps and to better support agent navigation.

% This approach should be broadly generalizable. For example, a similar approach, using the pre-trained GPT-3 could be used to explore other belief spaces, such as conspiracy theories, game strategies, or economic theories. These visualizations could be used as a way to explore the space of possibilities or as a way to map out the causal structure of population-scale behavior.

We strongly believe that this approach is generalizable, and can be applied in similar form to the narrative spaces that make up other regions such as philosophy (as we have seen), but also conspiracy theories, game strategies, etc. The potential application of creating graphical representations of the mental maps that exist in these topics is vast, and the methods employed here could be used to explore any complex topic.

% We would also like to see how these visualizations might be used in an educational context. The ability to see how one's beliefs about a domain change over time could be useful for improving one's understanding of a domain. For example, a student might be given a set of information and then be asked to construct a map that represents beliefs about that information. As the student leans more and adds topics and concepts to the map, they could watch how the space changes over time. We believe that this kind of visualization could provide a powerful way to improve one's understanding.

Finally, while this technique is generalizable, our example was a military one. While it has become rare for academia to contribute directly to an understanding of military thinking, now that AI is being actively utilized by armed forces, academia must participate vigorously in discussions about the ethical use of such technologies, as they possess a vital perspective into the risks inherent in these emerging technologies.

\section{Conclusions}
In this paper, we have shown that creating \textit{neural narrative maps} created from the output of a language model can be leveraged to create new meaningful information relationships. This process can be performed automatically if there is a source of ground truth, or iteratively, using direct human involvement to vet and connect concepts. This hybrid approach is flexible and allows humans to work on a more subjective level, filtering and directing the GPT-3 to articulate narratives that can be used to generate these visualizations. Our results show that we can use this process to create preliminary maps that are designed for human consumption, and we explore how these maps can be used to visualize the mental models of individuals or groups as they interact over time.

By combining topic extraction, machine learning, and human feedback, we can produce outputs that are both useful and understandable. These map-like representations can be used to explore beliefs, strategies, or even just preferences. We hope that this work will help others to visually explore and represent mental models as we work towards maps that can augment and support ethical decision making.

The role of usability is also important. Although a simple implementation, the ability of motion to attract attention the the relevant components on the map was substantial. We believe the incorporation of dynamic elements in these presentations will substantially improve the human comprehension of these maps.

We believe that this research is an important step toward creating automated tools that allows us to see relationships \textit{at scale}, between narrative elements that are otherwise hard to visualize and comprehend. We also demonstrated ways graphical representations of mental maps can be used for understanding how narratives are linked together.

To illustrate how this approach could impact tactical, strategic, and political thinking, we will consider another military scenario. In this case, it's a true story, about the price that can be paid for making ethical decisions.

On the night of July 27 2005, a group of four SEALs led by Lt. Michael Murphy were dropped into Afghanistan's Kunar province to set up an observation post. Around noon on the next day, two Afghan men and a 14-year-old boy with their small flock of goats stumbled on the post. The SEALs argued among themselves as to whether they should kill the civilians to protect their cover, detain them, or let them go and abandon the mission. In the end they decided that the right thing to do was to let the Afghans go and move the observation post. 

Before they had time to reposition, a force of nearly 100 Taliban fighters descended on their location from the same direction that the shepherds had fled. In the fierce firefight that followed, three of the four members of the team were killed, along with Lt. Murphy, while calling for support. The sole survivor, Marcus Luttrell, was rescued by local Pashtuns while escaping after being wounded by jumping off a series of cliffs~\cite{lucas2009not}.

In this case, Lt. Murphy made the \textit{ethically correct choice}\footnote{For which he was awarded a posthumous Medal of Honor.}.  Tragically, his death may have resulted from that same choice. However, the framing of the entire mission, where a tiny team was placed deep into a high-risk, poorly-understood region plays into the result as well. And at a still higher level, the abandonment of Afghanistan to the Taliban shows that many of the decisions made in that 20-year campaign were deeply flawed.

\begin{figure}[!htbp]
	\centering
	\fbox{\includegraphics[width=25em]{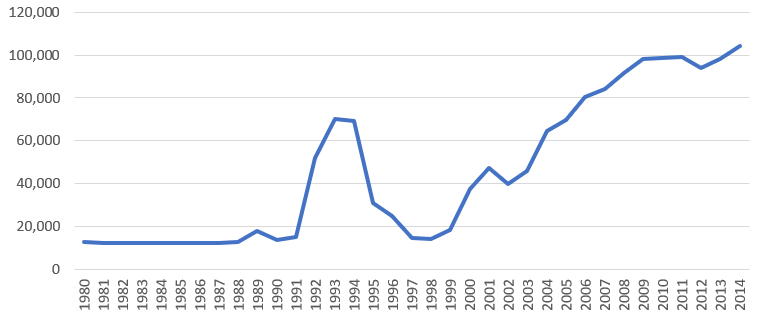}}
	\caption{\label{fig:UN_peackeeping} Size of total United Nations Peacekeeping Force (1980 - 2014)}
\end{figure}

Human beings have many biases. The more obvious involve gender, ethnicity, and race. But we also have subtle biases that affect how we make decisions on issues such as national security. For example, the USA has a bias towards advanced weapons systems~\cite{USA_Military_budget_2021}. This is reflected in the decisions to incorporate AI/ML into the nation's military. The focus is on intelligent munitions, drones, hypersonic missiles, etc. But since the end of the Cold War, the majority of military operations have been in irregular conflict, such as Kosovo, Libya, and Afghanistan. These conflicts often involve the United Nations in peacekeeping operations, and the presence of UN troops is an excellent proxy for the increase in irregular conflict~\cite{owidpeacekeeping} (Figure~\ref{fig:UN_peackeeping}). An intelligent munition would not have helped Lt. Murphy's team decide whether to kill, hold, or release the Afghan shepherds that stumbled upon them. But information presented in a way that lets a user clearly visualize the likely outcome of a \textit{trajectory of choices}, may let people consider other paths. After Vietnam, Iraq, and Afghanistan, leaders might think twice if that they see they are heading towards the part of a neural narrative map marked \enquote{Quagmire}. That would be a true ethical impact of AI in political and military thinking.
\section{Acknowledgements}

The authors would like to thank the \textit{Lockheed-Martin Artificial Intelligence Center} (LAIC) for funding a substantial part of the development of the interactive tools for subjective analysis. We would also like to acknowledge \emph{OpenAI}, whose GPT-3 provided the data for this research. Lastly, we would like to thank the reviewers, whose comments and suggestions dramatically improved this paper.


%\bibliographystyle{IEEEtran}
%\bibliography{biblio}


%\input{text/biographies}
%\input{text/author_instructions}
%\input{text/response}

\end{document}